\documentclass[11pt]{article}

\usepackage[utf8]{inputenc} 
\usepackage[T1]{fontenc}    
\usepackage[english]{babel}
\usepackage[margin=1in]{geometry}
\usepackage{amsmath}
\usepackage{amssymb}
\usepackage{amsfonts}
\usepackage{amsthm}
\usepackage{mathtools}
\usepackage{booktabs}       
\usepackage{nicefrac}       
\usepackage{microtype}      
\usepackage[table]{xcolor}  
\usepackage{graphicx}
\usepackage{subcaption}
\usepackage{float}
\usepackage{xspace}
\usepackage[varqu,varl,var0,scaled=0.97]{inconsolata}
\usepackage{listings}
\usepackage{tabularx}
\usepackage{makecell}
\usepackage[export]{adjustbox}
\usepackage{url}
\usepackage[round,authoryear]{natbib}
\usepackage{hyperref}
\usepackage[capitalize,noabbrev]{cleveref}

\usepackage{amsmath,amsfonts,bm}

\def\Figref#1{Fig.~\ref{#1}}

\def\eqref#1{equation~\ref{#1}}
\def\Eqref#1{Eq.~\ref{#1}}
\def\Tabref#1{Tab.~\ref{#1}}

\def\1{\bm{1}}

\DeclareMathAlphabet{\mathsfit}{\encodingdefault}{\sfdefault}{m}{sl}
\SetMathAlphabet{\mathsfit}{bold}{\encodingdefault}{\sfdefault}{bx}{n}

\theoremstyle{plain}

\theoremstyle{definition}

\theoremstyle{remark}

\newcommand{\dlr}{\textsc{DLR}\xspace}

\newcommand{\eg}{\emph{e.g.},\xspace}

\newcommand\fakeparagraph[1]{\par\noindent\textbf{{#1}}.\xspace}

\definecolor{dkgreen}{rgb}{0,0.6,0}
\definecolor{gray}{rgb}{0.5,0.5,0.5}
\definecolor{mauve}{rgb}{0.58,0,0.82}

\title{\textbf{DLR: Zero-Inference-Cost Latent Residuals for\\Low-Rank Pre-Training}}

\author{
Dong Wang\textsuperscript{$\diamond$}, 
Wenwu Tang\textsuperscript{$\diamond$},
Yun Cheng\textsuperscript{\textdagger}, 
Olga Saukh\textsuperscript{$\diamond$} \\
\textsuperscript{$\diamond$}Graz University of Technology, Austria \\
\textsuperscript{\textdagger}Swiss Data Science Center, Switzerland \\
\texttt{\{dong.wang, wenwu.tang, saukh\}@tugraz.at}, 
\texttt{yun.cheng@sdsc.ethz.ch} \\
}

\date{}

\begin{document}

\maketitle

\begin{abstract}
Large language models have driven recent progress in language and multimodal AI, yet pre-training them at scale is prohibitively expensive. Low-rank pre-training, which factorizes each weight matrix into a rank-$r$ product to reduce both parameters and FLOPs, is a promising response but typically lags full-rank training in quality. We propose Duplicated Latent Residual (\dlr), a training-only, parameter-free, foldable plug-in for low-rank pre-training. \dlr augments the standard low-rank output $Bz$ with a fixed structured residual $\tfrac{\alpha}{\sqrt K}\,\mathrm{Expand}_K(z)$ that replicates each latent coordinate $K{=}\lceil d_{\mathrm{out}}/r\rceil$ times across the output. With $\alpha$ fixed, \dlr adds zero learnable parameters per layer; after training, it is absorbed into the up-projection in closed form, $B^{\star} = B + \tfrac{\alpha}{\sqrt K}R^{\top}$, so deployment parameter count, FLOPs and memory match the underlying low-rank backbone exactly. Across LLaMA models from 60M to 7B parameters, \dlr strengthens low-rank pre-training on C4 validation perplexity in most settings, with the clearest gains at 130M and above; folded checkpoints transfer cleanly to supervised fine-tuning on standard benchmarks.
\end{abstract}

\section{Introduction}
\label{sec:intro}

Large language models (LLMs) have transformed machine learning across language, vision, and multimodal domains, yet this success comes at an unsustainable cost.
Pre-training modern LLMs requires vast compute, memory, and energy resources, with models like LLaMA-3~\citep{llama3}, Qwen-3~\citep{yang2025qwen3}, DeepSeek-V3~\citep{liu2024deepseek}, and GPT-4~\citep{gpt4} consuming thousands of GPU-months.
This has spurred significant research into efficient pre-training methods that preserve model quality while reducing training cost.

\begin{figure}[t]
    \centering
    \includegraphics[width=\linewidth,valign=t]{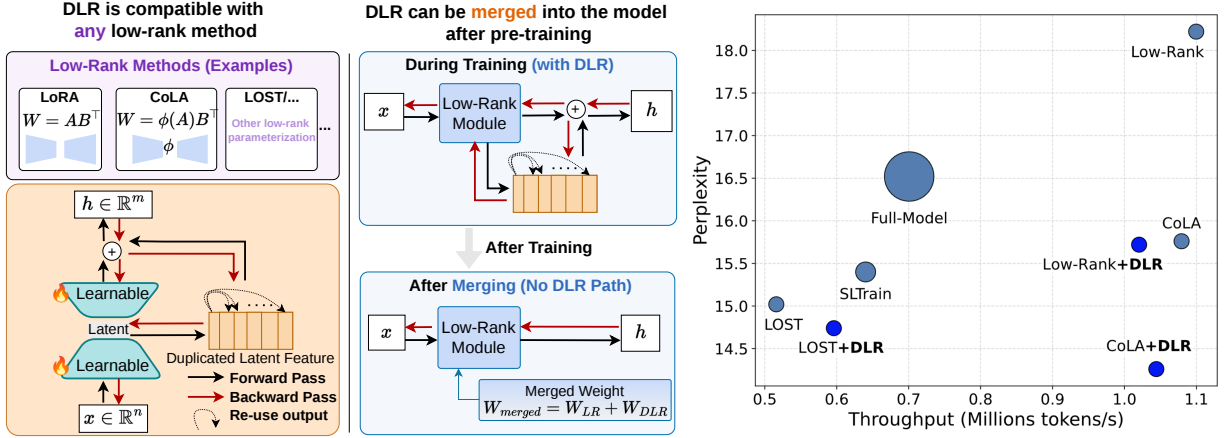}
    \caption{\textbf{Overview and perplexity--throughput trade-off.}
    (a) Duplicated Latent Residual (\dlr) is a training-only residual branch for low-rank decoders that is removed by folding before inference.
    (b) \dlr improves the perplexity--throughput trade-off at LLaMA-1B; full results are in \Tabref{tab:results}.}
    \label{fig:intro_overview}
\end{figure}

Existing efficient pre-training methods take three forms: extending LoRA-style low-rank updates to pre-training~\citep{relora,switchlora,loro}, projecting gradients into low-rank subspaces to compress optimizer states~\citep{gaLore,fira}, and modifying the network structure with low-rank parameterization possibly augmented by sparse compensation or latent-space mixing~\citep{sltrain,lost,cola,LaX}.
The third line is the most direct route to reducing both trainable parameters and forward FLOPs, but many additional pathways introduced to compensate for the rank-$r$ bottleneck (sparse residuals, channel-sparse compensation, or inter-layer latent gates) \emph{persist at inference}.
Deploying these augmented variants therefore requires shipping a modified inference graph rather than the vanilla rank-$r$ decoder, forfeiting one of low-rank pre-training's most attractive practical properties: drop-in compatibility with the existing low-rank inference stack.

A natural question arises: \emph{can we improve the training dynamics of low-rank backbones without sacrificing their parameter and compute efficiency?}

We propose \textbf{Duplicated Latent Residual} (\dlr), a \emph{training-only}, \emph{parameter-free}, \emph{foldable} plug-in for low-rank pre-training.
\dlr augments the standard low-rank output $Bz$ with a fixed structured residual that replicates each latent coordinate $K = \lceil d_{\mathrm{out}}/r \rceil$ times across the output dimension and rescales it by $\alpha/\sqrt{K}$, providing an additional gradient pathway that does not depend on the learned decoder (\Figref{fig:intro_overview}).
With $\alpha$ held fixed, no learnable parameters are added per layer at training time.
After training, the residual is absorbed into the up-projection in closed form via $B^{\star} = B + (\alpha/\sqrt{K})\,R^{\top}$, where $R \in \{0,1\}^{r \times d_{\mathrm{out}}}$ is the structured binary replication matrix induced by the duplication ($\mathrm{Expand}_K(z) = R^{\top} z$, see~\Eqref{eq:expand}).
The folded checkpoint runs on the underlying low-rank inference stack with the same parameter count, FLOPs, and memory footprint as the base backbone.

Empirically, \dlr strengthens four representative low-rank backbones (linear Low-Rank, CoLA~\citep{cola}, LOST~\citep{lost}, and LaX~\citep{LaX}) in most settings across LLaMA pre-training scales from 60M to 7B using a single untuned configuration, with mixed results only at the smallest 60M scale; the DLR+LOST combination at 1B improves PPL from 15.02 to 14.74, and in our single-seed 7B comparison DLR+CoLA outperforms reported LOST baselines from 40K steps onward while using $2.3{\times}$ less per-GPU memory (\Tabref{tab:results}, \Tabref{tab:results-7b}).
Folded checkpoints transfer cleanly to supervised fine-tuning on Alpaca-cleaned, where DLR+CoLA yields a slightly higher average accuracy among the three checkpoints we compare, despite deploying through the same inference graph as vanilla CoLA.

We summarize our contributions as follows:
\begin{itemize}
    \item \textbf{Foldable training residual.} We give a closed-form identity, $B^{\star} = B + \tfrac{\alpha}{\sqrt K}\,R^{\top}$, that absorbs \dlr into the up-projection of any low-rank parameterization that exposes a latent-to-output decoder, yielding the same deployed graph shape, parameter count, FLOPs, and memory footprint as the base method.
    \item \textbf{Parameter-free training overhead.} With $\alpha = 1$ fixed, \dlr adds zero learnable parameters per layer. Its training-time forward cost is a single \texttt{repeat\_interleave} plus add, with small overhead under \texttt{torch.compile} and zero inference overhead after folding.
    \item \textbf{Backbone-agnostic plug-in.} A single \dlr configuration improves four low-rank backbones (linear Low-Rank, CoLA, LOST, LaX) across LLaMA pre-training scales from 130M to 7B on C4 perplexity, with mixed results at 60M (\Tabref{tab:results}, \Tabref{tab:results-7b}); folded checkpoints transfer cleanly to Alpaca-cleaned SFT, where \dlr+ CoLA yields a slightly higher average accuracy among the compared 1B checkpoints (\Tabref{tab:sft}).
\end{itemize}
\section{Low-rank efficient pre-training}
\label{sec:method}

\subsection{Setup and a Unified View}
Consider a linear map in a transformer block (bias omitted):
\begin{equation}
    y = W x,\quad W\in\mathbb{R}^{d_{\mathrm{out}}\times d_{\mathrm{in}}},\ x\in\mathbb{R}^{d_{\mathrm{in}}}.
\end{equation}
Many efficient pre-training methods can be expressed as
\begin{equation}
\label{eq:unified}
    y \;=\; \underbrace{\Phi_{\mathrm{LR}}(x;\theta)}_{\text{low-rank backbone}}
    \;+\;
    \underbrace{\Phi_{\mathrm{EX}}(x;\psi)}_{\text{extra compensation}},
\end{equation}
where $\Phi_{\mathrm{LR}}$ is a low-rank backbone (possibly nonlinear in latent space),
and $\Phi_{\mathrm{EX}}$ is an additional compensation pathway.

We use a shared notation throughout. Let $r\ll \min(d_{\mathrm{in}},d_{\mathrm{out}})$ be the rank,
$A\in\mathbb{R}^{d_{\mathrm{in}}\times r}$ be the down-projection,
$B\in\mathbb{R}^{ d_{\mathrm{out}}\times r}$ be the up-projection, and define the latent
\begin{equation}
\label{eq:latent}
    z \;=\; s\cdot \phi(A^\top x)\in\mathbb{R}^{r},
\end{equation}
where $\phi(\cdot)$ is an optional activation and $s$ is a scalar scaling (fixed or learnable).
Unless stated otherwise, we focus on \emph{low-rank pre-training} where the linear map is parameterized by
$(A,B)$ (rather than an adapter added on top of a separate frozen base weight).

This unified decomposition allows us to compare seemingly different efficient pre-training methods through the lens of where and how they introduce additional learning signals beyond a rank-$r$ backbone. In particular, it makes clear whether improvements stem from modifying the latent representation or from adding new weight-space pathways.

\subsection{Baselines in the Unified Notation}

\fakeparagraph{Low-rank (linear)~\citep{lora}}
The linear low-rank parameterization replaces \(W\) by \(BA^\top\):
\begin{equation}
    \Phi_{\mathrm{LR}}(x)=B(A^\top x),\qquad \Phi_{\mathrm{EX}}(x)\equiv 0.
\end{equation}
This cuts multiplies from \(d_{\mathrm{out}}d_{\mathrm{in}}\) to \(r(d_{\mathrm{out}}{+}d_{\mathrm{in}})\) and reduces optimizer states to the size of \((A,B)\), but the fixed rank often underperforms full-rank pre-training~\citep{kamalakara2022exploring,khodak2021initialization}.

\fakeparagraph{SLTrain (low-rank + element-wise sparse residual)~\citep{sltrain}}
SLTrain augments the low-rank factorization with a fixed-support sparse residual:
\begin{equation}
\label{eq:sltrain}
    W = BA^\top + S,\qquad \mathrm{supp}(S)=\Omega,\ |\Omega|=k,
\end{equation}
where \(\Omega\subseteq[d_{\mathrm{out}}]\times[d_{\mathrm{in}}]\) is sampled once at initialization.
Equivalently,
\begin{equation}
    \Phi_{\mathrm{LR}}(x)=B (A^\top x),\qquad
    \Phi_{\mathrm{EX}}(x)=Sx.
\end{equation}
Thus SLTrain improves expressivity through an explicit sparse weight-space path, unlike \dlr's fixed latent-space residual that is folded into \(B\).

\fakeparagraph{LOST (low-rank + channel-sparse compensation)~\citep{lost}}
LOST combines a low-rank branch with an SVD-guided channel-sparse component intended to cover complementary directions.
\begin{equation}
\label{eq:lost}
    \Phi_{\mathrm{LR}}(x)=B z,\qquad
    \Phi_{\mathrm{EX}}(x)=W_s\,P_Ix,
\end{equation}
where \(P_I\in\{0,1\}^{k\times d_{\mathrm{in}}}\) selects a subset \(I\) of \(k\) input channels, so \(P_Ix\in\mathbb{R}^{k}\) contains exactly the channels feeding the learned compensation matrix \(W_s\in\mathbb{R}^{d_{\mathrm{out}}\times k}\); equivalently, the columns of \(W_s\) correspond to the \(P_I\)-selected channels.
The latent \(z\) is defined in~\Eqref{eq:latent}.
The compensation path is learned and remains part of the deployed graph, whereas \dlr adds a fixed latent residual during training and folds it into \(B\).

\fakeparagraph{CoLA (nonlinear latent low-rank)~\citep{cola}}
CoLA keeps a pure low-rank decoder but makes the latent representation nonlinear:
\begin{equation}
\label{eq:cola}
    \Phi_{\mathrm{LR}}(x)=B z,\qquad
    z=s\cdot\phi(A^\top x),\qquad
    \Phi_{\mathrm{EX}}(x)\equiv 0.
\end{equation}
It enriches trainability through latent gating without adding a separate weight-space compensation path, making it a natural backbone for \dlr.

\fakeparagraph{LaX (inter-layer latent residual)~\citep{LaX}}
LaX improves low-rank training by introducing an \emph{inter-layer} residual in latent space.
For a layer latent \(z_i=s\cdot\phi(A_i^\top x_i)\), it forms
\begin{equation}
    \tilde z_i=z_i+G(z_{i-1}),\qquad
    y_i=\mathrm{LN}(B_i\tilde z_i),
\end{equation}
where \(G\) is a lightweight alignment gate and the output LayerNorm follows the original formulation.
In the unified view, LaX keeps the same low-rank decoder form but adds information flow from the previous layer's latent, improving trainability without increasing the target rank \(r\).

Taken together, these methods illustrate two dominant strategies for improving low-rank pre-training: (i) adding auxiliary weight-space pathways to compensate for rank deficiency (\eg SLTrain, LOST), or (ii) enriching the latent representation while keeping a single low-rank decoder (\eg CoLA, LaX). \dlr is most naturally described as a latent-space plug-in, but it can attach to any backbone that exposes a low-rank latent-to-output decoder \(Bz\), including the low-rank branch of hybrid methods such as LOST. It therefore targets training dynamics through a fixed, non-learned expansion without tying the design to one particular low-rank backbone.

\section{Duplicated Latent Residual (\dlr)}
\label{sec:dlr}

\fakeparagraph{Key idea}
Given an existing low-rank layer with latent $z\in\mathbb{R}^r$ and output $Bz$,
\dlr attaches an \emph{intra-layer latent-space} residual that expands the same latent to output width
with a \emph{fixed structured decoder} as shown in \Figref{fig:intro_overview}.
Let $K=\lceil d_{\mathrm{out}}/r\rceil$ and define an expansion operator
\begin{equation}
\label{eq:expand}
    \mathrm{Expand}_K:\mathbb{R}^{r}\to\mathbb{R}^{d_{\mathrm{out}}},
    \;
    [\mathrm{Expand}_K(z)]_{i} = z_{\lfloor i/K\rfloor},
\end{equation}
with the last block truncated if $rK>d_{\mathrm{out}}$.
Equivalently, $\mathrm{Expand}_K(z)=R^\top z$ for a fixed binary replication matrix
$R\in\{0,1\}^{r\times d_{\mathrm{out}}}$ with $R_{j,i}=\mathbb{1}\{j=\lfloor i/K\rfloor\}$.

The plug-in form used during training is:
\begin{equation}
\label{eq:dlr}
    y \;=\; B z \;+\; \frac{\alpha}{\sqrt{K}}\cdot \mathrm{Expand}_K(z)\;+\;b,
\end{equation}
where $\alpha\in\mathbb{R}$ controls the residual strength (fixed in all experiments).
Duplicated Latent Residual (\dlr) therefore augments, rather than replaces, a standard low-rank layer: the backbone still supplies $A$, $B$, the activation, and any other latent-space structure, while \dlr contributes only the fixed residual term $\frac{\alpha}{\sqrt{K}}\,\mathrm{Expand}_K(z)$ during training.
Intuitively, \dlr views the output space as partially redundant and leverages structured replication to expose each latent coordinate to multiple output channels. Unlike sparse or channel-selective compensation, the resulting residual is dense, fixed, and highly structured, leading to uniform coverage of the output space without introducing irregular sparsity. As a result, \dlr provides an additional learning signal that bypasses the conditioning of the learned decoder, while retaining the computational characteristics of the underlying low-rank backbone.

\fakeparagraph{Foldability: zero inference overhead}\label{para:foldability}
A defining property of \dlr is that, after training, the duplicated-residual branch can be \emph{exactly} absorbed into the up-projection $B$ in closed form, leaving the inference graph identical to the underlying low-rank decoder.
Recall from~\Eqref{eq:expand} that $R \in \{0,1\}^{r \times d_{\mathrm{out}}}$ with $\mathrm{Expand}_K(z) = R^{\top} z$, so $R^{\top} \in \{0,1\}^{d_{\mathrm{out}} \times r}$ has the same shape as $B$ (each column of $R^{\top}$ is the indicator of one latent group).
Substituting into~\Eqref{eq:dlr} and collecting linear terms in $z$ yields the algebraic identity
\begin{equation}
\label{eq:fold}
y \;=\; B z \;+\; \tfrac{\alpha}{\sqrt{K}}\,R^{\top} z \;+\; b
   \;=\; \underbrace{\Big(B + \tfrac{\alpha}{\sqrt{K}}\,R^{\top}\Big)}_{\displaystyle B^{\star}}\, z \;+\; b
   \;=\; B^{\star}\, z \;+\; b ,
\end{equation}
where $B^{\star} \in \mathbb{R}^{d_{\mathrm{out}} \times r}$ has \emph{the same shape} as the original up-projection $B$.
Foldability follows by construction: the operation $B \leftarrow B + (\alpha/\sqrt{K})\,R^{\top}$ is performed once at training termination, and the resulting checkpoint is a drop-in replacement that runs on any code path implementing standard rank-$r$ inference (the $\mathrm{Expand}_K$ branch, the buffer storing $R$, and the residual scale $\alpha$ are all discarded).
Hence the deployment artifact has the same graph structure, parameter count, FLOPs, and peak-memory footprint as the corresponding low-rank baseline, while still benefiting from the better-trained $B^{\star}$.
This contrasts with prior residual or auxiliary-pathway approaches over low-rank backbones (e.g., LoR2C~\citep{LoR2C}, ResLoRA~\citep{ResLoRA}, sparse-additive variants such as SLTrain~\citep{sltrain}, or inter-layer latent residuals such as LaX~\citep{LaX}): they retain auxiliary parameters or non-mergeable nonlinearities at inference time and therefore modify the deployed graph.
Appendix~\ref{appx:related} further situates \dlr among residual-style methods and clarifies why foldability differs from mergeable adapter residuals (\Tabref{tab:related_residual_methods}).
A reference \texttt{fold()} implementation in $\sim$10 lines is provided in Appendix~\ref{appx:dlr_reference_impl}; the operation is the only \dlr-specific code that runs in the deployment pipeline, after which inference is indistinguishable from a vanilla rank-$r$ decoder.

\fakeparagraph{Variance-preserving correction}
A naive duplication would amplify the residual energy by a factor of $K_j$ in the $j$-th block.
We therefore use the global scaling $\beta=\alpha/\sqrt{K}$ in~\Eqref{eq:dlr}.
When $d_{\mathrm{out}}=rK$ (all blocks have size $K$), this choice preserves per-latent energy:
$\|\beta\,\mathrm{Expand}_K(z)\|_2^2 = \alpha^2\|z\|_2^2$.
Equivalently, if $\mathrm{Var}(z_j)=\sigma_z^2$, then each duplicated coordinate has variance
$\mathrm{Var}(\beta z_j)=\alpha^2\sigma_z^2/K$, so the \emph{total} variance mass assigned to the block remains
$\alpha^2\sigma_z^2$.
If the last block is truncated ($K_r<K$), the same scaling yields a slightly smaller energy contribution for that block, which is benign and avoids data- or shape-dependent re-scaling.
This correction mirrors the variance-preserving ``repair'' principle used in folding/merging operations~\citep{wang2025modelfolding,saukh2026cut} and is essential for stable pre-training~\citep{jordan2022repair}.

\fakeparagraph{Duplication map}
\dlr employs a \emph{deterministic, uniform} duplication structure.
The expansion operator $\mathrm{Expand}_K$ partitions the $d_{\mathrm{out}}$ output coordinates into $r$ consecutive blocks of size $K$ (with the last block possibly truncated to $d_{\mathrm{out}} - (r-1)K$ elements).
Each block $\mathcal{G}_j = \{jK, jK{+}1, \ldots, \min((j{+}1)K{-}1, d_{\mathrm{out}}{-}1)\}$ receives a copy of the latent coordinate $z_j$.
Formally, the replication matrix $R\in\{0,1\}^{r\times d_{\mathrm{out}}}$ has row $j$ containing ones at columns $\mathcal{G}_j$ and zeros elsewhere.
This fixed, balanced structure ensures that each latent dimension contributes equally to the residual pathway and requires no hyperparameter tuning beyond the rank $r$.
We implement $\mathrm{Expand}_K(z)$ as a single \texttt{repeat\_interleave} followed by truncation; the \emph{contiguous} form is our hardware-efficient choice, since more complex fixed maps (\eg permuted masks or non-contiguous/grouped duplication) force indexed gather/scatter operations (e.g., \texttt{index\_select}) that cannot be fused as cleanly. We quantify this implementation choice via a random-duplication ablation in Sec.~\ref{sec:experiments}.

\fakeparagraph{Design rationale}
\dlr is built around two deliberately simple choices: a fixed latent-to-output duplication map and a shape-dependent scale \(\beta=\alpha/\sqrt{K}\).
The scale prevents the residual branch from growing merely because each latent coordinate is copied multiple times.
For the group \(\mathcal{G}_j\) that receives copies of \(z_j\), the duplicated residual block satisfies
\[
    \|\beta (R^\top z)_{\mathcal{G}_j}\|_2^2
    = \beta^2 K_j z_j^2
    = \alpha^2 \frac{K_j}{K} z_j^2,
\]
so when the group is full (\(K_j=K\)) and \(\alpha=1\), the residual assigns the block the same squared energy as the original latent coordinate.
This normalization is purely shape-based: it keeps the forward residual on the same scale as \(z\) without introducing data-dependent statistics, per-layer tuning, or additional learned parameters.

The same fixed branch also changes the backward path in a controlled way.
Let \(g_y=\partial\mathcal{L}/\partial y\in\mathbb{R}^{d_{\mathrm{out}}}\).
Differentiating \Eqref{eq:dlr} with respect to the latent variable gives
\begin{equation}
\label{eq:gradz}
g_z \;=\; \frac{\partial\mathcal{L}}{\partial z}
\;=\; B^\top g_y \;+\; \beta R g_y.
\end{equation}
The first term is the usual gradient route through the learned decoder \(B\), while the second term is a fixed, decoder-independent route that aggregates output gradients according to the same groups used in the forward duplication.
This is the mechanism we intend \dlr to provide: during training, the encoder \(A\) receives an additional calibrated signal that does not have to pass through the current state of \(B\); after training, the entire route is folded into \(B^\star\) by \Eqref{eq:fold}.
The effectiveness of this design is evaluated empirically in the gradient-path measurements of Sec.~\ref{sec:experiments}, where the residual-induced component is strong early in training and remains nearly orthogonal to the learned-decoder component, and in the variance-correction ablation of \Tabref{tab:dlr_varcorr_1b}.

\fakeparagraph{Computational complexity}
\dlr does not replace the low-rank backbone; it adds a training-time residual branch on top of the backbone's existing latent $z$ and up-projection $B$. For any backbone that outputs $Bz$, the only extra forward work during training is the structured expansion-and-add term $\frac{\alpha}{\sqrt{K}}\,R^\top z$ (equivalently $\frac{\alpha}{\sqrt{K}}\,\mathrm{Expand}_K(z)$), implemented as \texttt{repeat\_interleave} followed by truncation (Appendix~\ref{appx:implementation}, Listing~\ref{lst:dlr_reference}). Crucially, the plug-in introduces no extra large GEMMs, no learned matrices, and no sparse/indexed matrix multiplications as in \Eqref{eq:sltrain}--\Eqref{eq:lost}; under \texttt{torch.compile}, our profiler traces do not show the expansion as a separate \texttt{aten::repeat\_interleave} kernel, suggesting fusion into surrounding computation. After folding, even this residual branch is removed from the inference graph; end-to-end throughput and peak memory are reported in Sec.~\ref{sec:experiments} (\Tabref{tab:system_1b_compile}).

\section{Experiments}
\label{sec:experiments}

\subsection{Evaluation Setup and Protocol}

\fakeparagraph{Pre-training data}
We pre-train all models on the Colossal Clean Crawled Corpus (C4)~\citep{raffel2020exploring}, a cleaned and deduplicated snapshot of Common Crawl that is widely used for language modeling.

\fakeparagraph{Model family and token budgets}
Following the experimental protocols used by \citet{sltrain}, \citet{glentis2025scalableparametermemoryefficient}, and \citet{lost}, we adopt the same LLaMA-style architecture family and match the token budget for each model scale. Concretely, we pre-train models with 60M, 130M, 350M, 1B, and 7B parameters~\citep{touvron2023llama2openfoundation,touvron2023llamaopenefficientfoundation}, using the same number of training tokens as prior work at each size (see \Tabref{tab:results}). Training details are provided in Appendix~\ref{appx:implementation}.

\fakeparagraph{Baselines}
We compare against \emph{Full-Rank} training and representative efficient pre-training methods, including
\emph{Low-rank}~\citep{lora}
\emph{GaLore}~\citep{gaLore},
\emph{Fira}~\citep{fira},
\emph{LORO}~\citep{loro},
\emph{SLTrain}~\citep{sltrain},
\emph{LOST}~\citep{lost},
\emph{CoLA}~\citep{cola}, and
\emph{LaX}~\citep{LaX},

Unless otherwise stated, all baselines use the authors' recommended hyperparameters and the same optimizer/schedule.

\fakeparagraph{Plugging \dlr into LLaMA backbones}
For each low-rank LLaMA backbone, we attach the \dlr residual to every low-rank linear projection in the Transformer blocks, including self-attention projections ($W_Q$, $W_K$, $W_V$, $W_O$) and the MLP/FFN projections. The backbone still determines the learned matrices, rank, activation, and any inter-layer latent structure. \dlr only adds the fixed duplicated-latent branch during training and folds it into the corresponding up-projection before deployment (Sec.~\ref{sec:dlr}).

To ensure a fair comparison, we use the same AdamW optimizer and closely follow the training recipe in prior work, including learning-rate schedule, warmup ratio, and packed-data training, whenever applicable~\citep{gaLore,glentis2025scalableparametermemoryefficient}. Additional implementation details (including rank settings, scaling choices, and compilation/profiling configurations) are provided in Appendix~\ref{appx:implementation}.

\subsection{End-to-End Performance}

\fakeparagraph{Main results}
We focus on the trade-off between validation perplexity (PPL) and training efficiency under matched token budgets. Improvements are meaningful only if they preserve the throughput and memory profile of the underlying low-rank backbone.
\Tabref{tab:results} reports validation PPL on C4 across LLaMA scales 60M / 130M / 350M / 1B, together with parameter count and estimated memory, paired by backbone (Low-Rank, CoLA, LOST, Low-Rank$+$LaX, CoLA$+$LaX) so that each base method can be read off against its $+$DLR counterpart.
A single untuned \dlr configuration ($\alpha{=}1$, fixed, uniform $\mathrm{Expand}_K$) consistently improves all four backbones at 130M and above while keeping the same parameter budget as the corresponding backbone.
In particular, CoLA$+$\dlr achieves substantial PPL reductions over CoLA at fixed rank (\eg $15.76{\rightarrow}14.26$ at 1B; $25.61{\rightarrow}23.80$ at 130M), and \dlr also improves the strong LOST baseline (\eg $15.02{\rightarrow}14.74$ at 1B), yielding the best perplexity--efficiency trade-off among the foldable plug-in variants we report.

\begin{table}[t]
    \centering
    \vspace{0.1in}
    \caption{\textbf{Main C4 validation results across LLaMA scales.}
    We report validation perplexity (PPL), parameter count (M), and estimated optimizer-state memory (GB) for 60M--1B models under matched token budgets.
    Rows are paired by backbone; \textbf{bold} marks the better PPL within each base/\(+\)\dlr pair.
    \dlr adds no learned parameters and folds into the base decoder, so each \(+\)\dlr row has the same deployed parameter count, FLOPs, and memory footprint as its base row.
    Non-\dlr baselines are reproduced by us or taken from prior work~\citep{glentis2025scalableparametermemoryefficient,lost,loro,sltrain,LaX,cola,gaLore,fira}; memory follows the optimizer-state estimate convention of~\citep{gaLore,sltrain}.}
    \vspace{0.1in}
    \renewcommand{\arraystretch}{1.2}
    \setlength{\tabcolsep}{3pt}
    \resizebox{0.99\textwidth}{!}{%
    \begin{tabular}{l|ccc|ccc|ccc|ccc}
    \toprule
    & \multicolumn{3}{c|}{\textbf{60M}} & \multicolumn{3}{c|}{\textbf{130M}} & \multicolumn{3}{c|}{\textbf{350M}} & \multicolumn{3}{c}{\textbf{1B}} \\
    {$r~/~d$}
        & \multicolumn{3}{c|}{128 / 512}
        & \multicolumn{3}{c|}{256 / 768}
        & \multicolumn{3}{c|}{256 / 1024}
        & \multicolumn{3}{c}{512 / 2048} \\
    {Tokens}
        & \multicolumn{3}{c|}{1.4B}
        & \multicolumn{3}{c|}{2.6B}
        & \multicolumn{3}{c|}{7.8B}
        & \multicolumn{3}{c}{13.1B} \\
    \midrule
    & PPL & Param & Mem & PPL & Param & Mem & PPL & Param & Mem & PPL & Param & Mem \\
    \midrule
    Full-Model        & 30.27 &   58 & 0.35 & 23.13 &  134 & 0.81 & 18.76 &  368 & 2.21 & 13.40 & 1339 & 8.04 \\
    LoRA~\citep{lora}         & 35.30 &   43 & 0.36 & 25.07 &   94 & 0.84 & 19.13 &  185 & 1.85 & 15.83 &  609 & 6.34 \\
    GaLore~\citep{gaLore}     & 34.58 &   58 & 0.28 & 25.31 &  134 & 0.61 & 19.37 &  368 & 1.59 & 15.57 & 1339 & 4.76 \\
    Fira~\citep{fira}         & 30.34 &   58 & 0.28 & 22.96 &  134 & 0.61 & 16.82 &  368 & 1.59 & 15.10 & 1339 & 4.76 \\
    SLTrain~\citep{sltrain}   & 32.58 &   47 & 0.30 & 24.17 &  104 & 0.67 & 18.59 &  215 & 1.54 & 15.40 &  732 & 5.33 \\
    LORO~\citep{loro}         & 33.87 &   43 & 0.24 & 24.78 &   94 & 0.57 & 19.66 &  185 & 1.11 & 15.53 &  609 & 3.66 \\
    \midrule
    Low-Rank                 & 35.13 &   43 & 0.24 & 26.71 &   94 & 0.57 & 21.77 &  185 & 1.11 & 18.22 &  609 & 3.66 \\
    \;$+$\textbf{DLR}        & \textbf{35.01}\textsubscript{\tiny$\pm$0.18} &   43 & 0.24
                             & \textbf{25.00}\textsubscript{\tiny$\pm$0.35} &   94 & 0.57
                             & \textbf{18.75}\textsubscript{\tiny$\pm$0.03} &  185 & 1.11
                             & \textbf{15.72}\textsubscript{\tiny$\pm$0.74} &  609 & 3.66 \\
    \midrule
    CoLA~\citep{cola}         & 34.10 &   43 & 0.24 & 25.61 &   94 & 0.57 & 19.75 &  185 & 1.11 & 15.76 &  609 & 3.66 \\
    \;$+$\textbf{DLR}        & \textbf{32.96}\textsubscript{\tiny$\pm$0.06} &   43 & 0.24
                             & \textbf{23.80}\textsubscript{\tiny$\pm$0.40} &   94 & 0.57
                             & \textbf{18.38}\textsubscript{\tiny$\pm$0.03} &  185 & 1.11
                             & \textbf{14.26}\textsubscript{\tiny$\pm$0.02} &  609 & 3.66 \\
    \midrule
    LOST~\citep{lost}         & \textbf{32.25} &   43 & 0.24 & 24.05 &   94 & 0.57 & 18.95 &  185 & 1.11 & 15.02 &  609 & 3.66 \\
    \;$+$\textbf{DLR}        & 33.15\textsubscript{\tiny$\pm$0.12}          &   43 & 0.24
                             & \textbf{24.03}\textsubscript{\tiny$\pm$0.03} &   94 & 0.57
                             & \textbf{18.88}\textsubscript{\tiny$\pm$0.02} &  185 & 1.11
                             & \textbf{14.74}\textsubscript{\tiny$\pm$0.01} &  609 & 3.66 \\
    \midrule
    Low-Rank$+$LaX~\citep{LaX} & \textbf{33.54} &   44 & 0.33 & 24.63 &   94 & 0.70 & 18.90 &  185 & 1.38 & 15.51 &  609 & 4.54 \\
    \;$+$\textbf{DLR}        & 33.71\textsubscript{\tiny$\pm$0.05}          &   44 & 0.33
                             & \textbf{24.38}\textsubscript{\tiny$\pm$0.02} &   94 & 0.70
                             & \textbf{18.85}\textsubscript{\tiny$\pm$0.01} &  185 & 1.38
                             & \textbf{15.29}\textsubscript{\tiny$\pm$0.48} &  609 & 4.54 \\
    CoLA$+$LaX~\citep{LaX}    & \textbf{33.21} &   44 & 0.33 & 24.21 &   99 & 0.74 & 18.51 &  196 & 1.46 & 14.78 &  609 & 4.54 \\
    \;$+$\textbf{DLR}        & 34.89\textsubscript{\tiny$\pm$0.07}          &   44 & 0.33
                             & \textbf{24.08}\textsubscript{\tiny$\pm$0.07}          &   99 & 0.74
                             & \textbf{18.37}\textsubscript{\tiny$\pm$0.02} &  196 & 1.46
                             & \textbf{14.67}\textsubscript{\tiny$\pm$0.00} &  609 & 4.54 \\
    \bottomrule
    \end{tabular}
    }
    \label{tab:results}
\end{table}

The 60M scale shows a few sub-cells (LOST$+$\dlr, Low-Rank$+$LaX$+$\dlr, CoLA$+$LaX$+$\dlr) where the $+$DLR row matches or slightly trails the corresponding base baseline.
We attribute this to multi-seed noise at the smallest scale: with the lowest token budget in our study (1.4B vs.\ 13.1B at 1B) and the smallest hidden width, each backbone's run-to-run PPL spread is wider relative to the absolute PPL gap a single-layer plug-in can move. The $\pm$std reported on the $+$DLR rows ($0.05$--$0.18$) is consistent with this regime, and the deviations vanish from 130M onward where the token-to-parameter ratio is larger.
We therefore do not draw conclusions from the 60M sub-cells in isolation, and report 60M primarily as the small end of a 4-scale trajectory rather than as a stand-alone benchmark.

Scaling-up results at 7B are reported separately in \Tabref{tab:results-7b}, and downstream results after instruction fine-tuning are presented in \Cref{sec:sft} (\Tabref{tab:sft}).
We additionally report pre-fine-tuning zero-shot evaluation for the 1B models in Appendix~\ref{appx:zero_shot_eval} (\Tabref{tab:zeroshot_1b}).

\fakeparagraph{Variance/scale correction (\(1/\sqrt{K}\))}
The duplicated residual copies each latent coordinate into \(K\) output coordinates, so we scale it by \(1/\sqrt K\) to keep the residual energy comparable across ranks and output widths.
Ablating this correction consistently worsens validation perplexity for DLR+CoLA across scales, indicating that the normalization is important in our pre-training setting.
Full results are reported in Appendix~\ref{appx:additional_ablation} (\Tabref{tab:dlr_varcorr_1b}).

\fakeparagraph{Effective-rank / target-rank sensitivity}
We also test whether \dlr remains useful under tighter rank budgets by reducing the default rank to \(0.75r_0\) and \(0.5r_0\).
Perplexity degrades smoothly rather than catastrophically, indicating a predictable quality--efficiency trade-off rather than dependence on a single tuned rank.
The full rank-sensitivity table is in Appendix~\ref{appx:additional_ablation} (\Tabref{tab:dlr_rank_sensitivity}).

\fakeparagraph{Scaling-law token budget (1B)}
To check whether the 1B gains persist beyond the standard 13.1B-token budget, we additionally train the 1B setting for 26B tokens.
The low-rank backbone equipped with \dlr continues to closely track the full-rank trajectory throughout training, with a small but consistent convergence gap, matching the final ordering in \Tabref{tab:results}.
The full trajectory is shown in Appendix~\ref{appx:additional_ablation} (\Figref{fig:eval_loss_curve_1b_26b}).

\fakeparagraph{Gradient-path analysis (LLaMA-350M)}
To understand why \dlr helps low-rank pre-training, we decompose the latent gradient into the learned-decoder path \(B^\top g_y\) and the duplicated-residual path \(\beta R g_y\), where \(\beta=\alpha/\sqrt{K}\) (cf.~\Eqref{eq:gradz}).
During LLaMA-350M pre-training, we track their relative magnitude \(\rho(t)=\|\beta R g_y\|/\|B^\top g_y\|\) and directional alignment \(\cos_t\) across representative projections (details in Appendix~\ref{app:gradlog}).
The residual-induced component is strongest early in training and then decays as the learned decoder takes over, while its cosine with the decoder path remains close to zero.
This supports the view that \dlr improves optimization by adding a calibrated, complementary gradient route rather than merely rescaling the existing low-rank update.
A complementary initialization-offset ablation in Appendix~\ref{appx:init_offset_ablation} separates this active training-time effect from simply initializing \(B\) with the folded block-constant offset.

\fakeparagraph{Choosing the residual scale \(\alpha\)}
Our formulation allows at most one scalar \(\alpha\) per layer to control the strength of the duplicated-latent residual in~\Eqref{eq:dlr}.
In practice, however, we find that \(\alpha\) is not a sensitive knob in our pre-training setting: fixing \(\alpha{=}1.0\) yields essentially identical training dynamics and final perplexity compared to making \(\alpha\) learnable, and a coarse hyperparameter sweep consistently selects values near \(\alpha{=}1.0\) as optimal.
We therefore use a fixed, non-trainable \(\alpha{=}1.0\) in all experiments, which keeps the plug-in parameter-free in practice.

\fakeparagraph{The role of $B$ and why we use a fixed contiguous duplication}
We ablate two design choices on LLaMA-60M: whether the learnable decoder \(B\) remains necessary, and whether the fixed duplication must be implemented as a contiguous \(\mathrm{Expand}_K\).
Removing \(B\) substantially degrades convergence, showing that \dlr is a complement to the learned low-rank decoder rather than a replacement for it.
Replacing contiguous duplication with a fixed random per-output mapping reaches similar loss but is much slower under \texttt{torch.compile}, because indexing-based gathers do not fuse as cleanly as \texttt{repeat\_interleave}.
Thus the default design keeps \(B\), uses a fixed contiguous map, and remains both foldable at inference and cheap during training; details are in Appendix~\ref{app:dlr-dynamics-60m}.

\fakeparagraph{Scaling to LLaMA-7B: DLR+CoLA enables low-memory pre-training at scale}
To probe scaling beyond 1B, we attach \dlr to a CoLA LLaMA-7B backbone and compare against Full-Rank Adam, 8-bit SLTrain, and LOST on the same C4 schedule (\Tabref{tab:results-7b}).
Baselines are taken from prior work~\citep{sltrain,lost,glentis2025scalableparametermemoryefficient}. Our DLR+CoLA run uses the matched protocol described here (rank $r{=}1024$, batch size 8, BF16, Adam states, single seed).
From 40K steps onward in this comparison, DLR+CoLA achieves the lowest perplexity at every milestone (12.08 at 150K vs.\ LOST 12.80) while using $2.3{\times}$ less per-GPU memory than LOST (27.03GB vs.\ 62.15GB). Full-Rank Adam and 8-bit SLTrain encounter OOM at this batch size.
This shows that the foldable plug-in extends to a memory-bound 7B setting while achieving the best perplexity among the compared runs.

\begin{table}[H]
\centering
\caption{\textbf{LLaMA-7B pre-training on C4.}
Validation perplexity at 10K--150K steps and steady-state per-GPU peak memory.
Baselines are reported from~\citep{sltrain,lost,glentis2025scalableparametermemoryefficient}; DLR+CoLA uses the matched protocol described in this paper.
``OOM'' marks runs that cannot advance at the listed batch size; \textbf{bold} indicates the best PPL at each milestone.}
\label{tab:results-7b}
\vspace{0.05in}
\small
\setlength{\tabcolsep}{4.5pt}
\begin{tabular}{lcc|ccccc}
\toprule
Method & Batch & Mem (GB) & 10K & 40K & 80K & 120K & 150K \\
\midrule
Full-Rank Adam & 4 & 49.53 & 24.95 & 20.05 & --- & OOM   & --- \\
8-bit SLTrain  & 8 & 60.91 & 27.59 & ---   & OOM & ---   & --- \\
LOST           & 8 & 62.15 & \textbf{24.41} & 16.48 & 14.01 & 12.93 & 12.80 \\
\textbf{\dlr+CoLA (ours)} & 8 & \textbf{27.03} & 24.98 & \textbf{15.82} & \textbf{13.71} & \textbf{12.69} & \textbf{12.08} \\
\bottomrule
\end{tabular}
\end{table}

\fakeparagraph{System measurements (LLaMA 1B)}
To corroborate that \dlr improves perplexity without introducing a practical efficiency bottleneck, we report end-to-end system measurements on 1B models with \texttt{torch.compile} enabled in \Tabref{tab:system_1b_compile}.
Each entry reports validation PPL together with peak GPU memory after initialization and training throughput after warmup, under the same multi-node H100 setup described in the table caption.
On the nonlinear low-rank backbone, adding \dlr improves CoLA from PPL 15.76 to 14.26 while keeping memory and throughput nearly unchanged (13.10$\rightarrow$13.09GB; 1,079,350$\rightarrow$1,044,264 tok/s).
On the linear low-rank backbone, \dlr improves PPL 18.22$\rightarrow$15.72 with only a small memory increase (12.58$\rightarrow$13.09GB) and a modest throughput decrease (1,099,699$\rightarrow$1,020,316 tok/s).
In contrast, sparse-compensation baselines show substantially lower throughput (SLTrain 640,132 tok/s; LOST 515,971 tok/s) and/or higher peak memory (14.55GB and 19.33GB), highlighting that \dlr retains the efficiency profile of pure low-rank training while recovering much of the quality gap.

\begin{table}[h]
\caption{
System measurements on 1B models with \texttt{torch.compile} enabled.
All runs use DDP on 8 nodes $\times$ 4 NVIDIA H100 SXM5 GPUs (94GB per GPU),
a local 7.68TB NVMe drive, and 4$\times$ Infiniband NDR200.
We report validation PPL, total parameter count, max GPU memory after initialization
(\texttt{max\_memory}), and throughput (global tokens/s) after warmup.
}
\label{tab:system_1b_compile}
\vspace{0.1in}
\centering
\begin{small}
\begin{tabular}{lcccc}
\toprule
Method &
\multicolumn{1}{c}{PPL} &
\multicolumn{1}{c}{Params} &
\multicolumn{1}{c}{Memory} &
\multicolumn{1}{c}{Throughput} \\
& $\downarrow$ & (M) & (GB) & (tok/s) \\
\midrule
Full-Rank            & 13.40 & 1339 & 17.45 & 700697 \\
Low-Rank             & 18.22 & 609  & 12.58 & 1099699 \\
\makecell[l]{Low-Rank + \dlr}        & 15.72 & 609  & 13.09 & 1020316 \\
\makecell[l]{Low-Rank + LaX + \dlr} & 15.29 & 610 & 14.96 & 878076 \\
CoLA                 & 15.76 & 609  & 13.10 & 1079350 \\
CoLA + \dlr            & 14.26 & 609  & 13.09 & 1044264 \\
\makecell[l]{CoLA + LaX}    & 14.78 & 610  & 15.58 & 846626 \\
\makecell[l]{CoLA + LaX + \dlr}          & 14.67 & 610  & 15.59 & 841800 \\
SLTrain              & 15.40 & 609  & 14.55 & 640132 \\
LOST                 & 15.02 & 609  & 19.33 & 515971 \\
\makecell[l]{LOST + \dlr}    & 14.74 & 609  & 19.33 & 595978 \\

\bottomrule
\end{tabular}
\end{small}
\end{table}

\subsection{Folded checkpoints transfer cleanly to supervised fine-tuning}
\label{sec:sft}

A core appeal of foldability is that the training-only residual leaves no trace at deployment.
We test this end-to-end by taking the 1B Full-Rank, CoLA, and DLR+CoLA pre-trained checkpoints, folding the \dlr branch into the up-projection (Sec.~\ref{sec:dlr}), running full-parameter SFT on Alpaca-cleaned, and evaluating six standard benchmarks with \texttt{lm-evaluation-harness}~\citep{eval-harness} (details in Appendix~\ref{appx:sft_details}).

\Tabref{tab:sft} reports mean $\pm$ std accuracy across three SFT seeds applied to the same fixed pre-trained checkpoint per method.
\dlr+CoLA attains a slightly higher average ($39.81\%$, vs.\ $39.55\%$ Full-Rank and $39.39\%$ CoLA), narrows the HellaSwag gap to Full-Rank, and achieves the highest PIQA accuracy.
This supports the \emph{train with \dlr $\rightarrow$ fold $\rightarrow$ deploy} workflow: the downstream gain appears after folding, while deployment keeps the same graph, parameter count, FLOPs, and memory footprint as vanilla CoLA.

\begin{table}[h]
\centering
\vspace{0.05in}
\caption{\textbf{Downstream evaluation after Alpaca-cleaned SFT (1B models).}
Each cell reports zero-shot accuracy as mean $\pm$ standard deviation across three SFT seeds applied to the same fixed pre-trained checkpoint.
\dlr+CoLA achieves a slightly higher average while folding to the same deployed graph as vanilla CoLA; \textbf{bold} indicates the best per-task mean.}
\label{tab:sft}
\vspace{0.05in}
\small
\setlength{\tabcolsep}{6pt}
\begin{tabular}{lccc}
\toprule
Task & \dlr+CoLA & CoLA & Full-Rank \\
\midrule
ARC-Challenge & 0.2327 $\pm$ 0.0027 & \textbf{0.2361 $\pm$ 0.0010} & 0.2190 $\pm$ 0.0013 \\
BoolQ & 0.3783 $\pm$ 0.0000 & 0.3783 $\pm$ 0.0000 & 0.3783 $\pm$ 0.0000 \\
HellaSwag & 0.3519 $\pm$ 0.0011 & 0.3296 $\pm$ 0.0009 & \textbf{0.3564 $\pm$ 0.0001} \\
MMLU & 0.2295 $\pm$ 0.0000 & 0.2295 $\pm$ 0.0000 & 0.2295 $\pm$ 0.0000 \\
PIQA & \textbf{0.6750 $\pm$ 0.0022} & 0.6649 $\pm$ 0.0005 & 0.6674 $\pm$ 0.0027 \\
WinoGrande & 0.5212 $\pm$ 0.0012 & \textbf{0.5254 $\pm$ 0.0061} & 0.5222 $\pm$ 0.0020 \\
\midrule
Average & \textbf{0.3981 $\pm$ 0.0004} & 0.3939 $\pm$ 0.0011 & 0.3955 $\pm$ 0.0006 \\
\bottomrule
\end{tabular}
\end{table}

\section{Conclusion, Limitations and Outlook}

We introduced \dlr, a training-only, parameter-free residual plug-in for low-rank pre-training that is folded into the up-projection after training, leaving the deployed low-rank graph unchanged.
Across LLaMA scales from 60M to 7B, \dlr improves representative low-rank backbones in most matched-token-budget settings, retains the efficiency profile of the base decoder, and transfers cleanly after supervised fine-tuning.
Together, these results support foldable latent residuals as a practical way to improve low-rank pre-training dynamics without adding inference-time complexity.

\textbf{Limitations.}
\dlr improves training dynamics but does not remove the approximation ceiling imposed by a fixed rank-$r$ bottleneck.
Its scale is shape-based and fixed at \(\alpha=1\) in our experiments. We do not yet characterize how to retune it for architectures or rank regimes with substantially different latent statistics.
Our evaluation focuses on LLaMA-style language models trained on C4 and instruction-tuned on Alpaca-cleaned, leaving other modalities, architectures, and tasks for future work.

\textbf{Outlook.}
More broadly, any training-time latent intervention that remains linear in \(z\) may admit a closed-form fold, suggesting a design space of inference-free training plug-ins beyond the particular contiguous duplication map studied here.

\section*{Acknowledgements}
This work has been supported by the FFG COMET K1 Center ``Pro\textsuperscript{2}Future II'' (Cognitive and Sustainable Products and Production Systems of the Future), Contract No.~911655.
The results presented in this paper were computed using the computational resources of Pro2Future GmbH, the Central IT Services of Graz University of Technology (ZID), and the Austrian Scientific Computing (ASC) infrastructure.

\newpage
\bibliographystyle{plainnat}
\bibliography{reference}
\newpage
\appendix
\section*{Appendix}

The following sections provide supplementary information omitted from the main text:
\begin{itemize}
    \item Appendix~\ref{appx:impact}: Impact statement and existing assets.
    \item Appendix~\ref{appx:implementation}: Implementation details.
    \item Appendix~\ref{appx:additional_ablation}: Additional ablations and long-horizon results.
    \item Appendix~\ref{app:gradlog}: Gradient-path analysis results.
    \item Appendix~\ref{app:dlr-dynamics-60m}: The impact of \dlr on training dynamics.
    \item Appendix~\ref{appx:sft_details}: Supervised fine-tuning details.
    \item Appendix~\ref{appx:zero_shot_eval}: Zero-shot evaluation.
    \item Appendix~\ref{appx:related}: Related work.
    \item Appendix~\ref{appx:use-of-llm}: Use of Large Language Models.
\end{itemize}

\section{Implementation details}
\label{appx:implementation}

We trained over 100 models to evaluate the performance of DLR presented in this work. Experiments were run on an NVIDIA H100 SLURM cluster where each node has 4 NVIDIA H100 GPUs (each with 94GB memory), a local 7.68TB NVMe drive, and 4$\times$ Infiniband NDR200 adapters. Training time varies from 20 minutes (LLaMA 60M) to 5 hours (LLaMA 1B) depending on the model size. \texttt{Huggingface Hub}\footnote{https://huggingface.co/docs/hub/index} is used to load the datasets. Weights \& Biases (W\&B)\footnote{https://wandb.ai} is used to log training history, training result, and evaluation metrics. The source code is available at \url{https://github.com/nanguoyu/DLR}.

This section outlines the LLaMA architectures and the hyperparameters used during pre-training. To ensure fair comparison, we follow the same experimental settings as \citet{gaLore,glentis2025scalableparametermemoryefficient}. \Tabref{tab:train_hparams} summarizes the hyperparameters for different model scales. Across all architectures, we adopt a maximum sequence length of 256 and a batch size of 131,072 tokens. The learning rate is linearly warmed up during the first 10\% of training steps, followed by a cosine annealing schedule that decays to 10\% of the initial value. We use the T5-base tokenizer~\citep{raffel2023t5}, consistent with prior work~\citep{sltrain,glentis2025scalableparametermemoryefficient}.
\begin{table}[h]
\centering
\caption{Pre-training hyperparameters for LLaMA architectures.}
\label{tab:pretrain-settings}
\vskip 0.1in
\renewcommand{\arraystretch}{1.15}
\setlength{\tabcolsep}{4pt}
\begin{small}
\begin{tabular}{@{}ccccccc@{}}
\toprule
Params & Hidden & Intermediate & Heads & Layers & Steps & Tokens (B) \\
\midrule
60M    & 512    & 1376         & 8     & 8      & 11K   & 1.4  \\
130M   & 768    & 2048         & 12    & 12     & 22K   & 2.6  \\
350M   & 1024   & 2736         & 16    & 24     & 65K   & 7.8  \\
1B     & 2048   & 5461         & 32    & 24     & 140K  & 13.1 \\
\bottomrule
\end{tabular}
\end{small}
\vskip -0.05in
\end{table}

\begin{table}[h]
\centering
\vspace{0.1in}
\small
\setlength{\tabcolsep}{3pt}
\renewcommand{\arraystretch}{1.2}
\caption{Training hyperparameters used in our DLR experiments (C4 pre-training). Common settings across these runs: C4 dataset; sequence length 256; token batch size $512\times256=131{,}072$ tokens; tokenizer T5-base; optimizer AdamW with $(\beta_1,\beta_2)=(0.9,0.999)$, weight decay $0.1$, and gradient clipping $0.5$; cosine learning-rate schedule with minimum LR ratio $0.1$; evaluation every 1{,}000 update steps; dtype bfloat16; seeds $\{41,42,43,44,45\}$.}
\vspace{0.1in}
\label{tab:train_hparams}
\begin{tabular}{lcccccccc}
\toprule
Model & World Size & Batch & Total Batch & $r$ & LR & Steps & Warmup & AdamW $\epsilon$ \\
\midrule
60M  & 8  & 64 & 512 & 128 & 0.01  & 11{,}000  & 1{,}100  & $10^{-8}$ \\
130M & 16 & 32 & 512 & 256 & 0.005 & 22{,}000  & 2{,}200  & $10^{-6}$ \\
350M & 16 & 32 & 512 & 256 & 0.003 & 65{,}000  & 6{,}500  & $10^{-6}$ \\
1B   & 32 & 16 & 512 & 512 & 0.002 & 140{,}000 & 10{,}000 & $10^{-6}$ \\
\bottomrule
\end{tabular}

\vspace{2pt}
\footnotesize
\noindent
\end{table}

\subsection*{Reference implementation of \dlr}
\label{appx:dlr_reference_impl}
To make the definition of the expansion operator \(\mathrm{Expand}_K(\cdot)\) concrete, we provide a short Python-style reference implementation of a \dlr layer below.
The paper writes the up-projection as \(B\in\mathbb{R}^{d_{\mathrm{out}}\times r}\) and \(Bz\); the code below uses the common PyTorch row-major convention \(B_{\mathrm{code}}\in\mathbb{R}^{r\times d_{\mathrm{out}}}\) and computes \texttt{z @ B}, which is the transpose convention of the mathematical \(B\).
The key operation is to replicate each latent coordinate \(K=\lceil d_{\text{out}}/r\rceil\) times along the last dimension and truncate to \(d_{\text{out}}\), matching \Eqref{eq:expand}--\Eqref{eq:dlr}.

\lstset{firstnumber=1}
\vspace{2pt}
\begin{lstlisting}[language=Python,caption={Python-style reference code for the \dlr layer (Expand via repeat+truncate).},label={lst:dlr_reference}]
import math

def expand_k(z, d_out):
    # z: [..., r]
    r = z.shape[-1]
    K = math.ceil(d_out / r)
    # Repeat each latent coordinate K times (e.g., PyTorch: z.repeat_interleave(K, dim=-1))
    u = repeat_interleave(z, K, axis=-1)   # length r*K
    return u[..., :d_out]                  # truncate to d_out

def dlr_layer(x, A, B, alpha, bias=None, act=None):
    # x: [..., d_in], A: [d_in, r], B: [r, d_out]
    # B is B_code = B_math^T, so z @ B corresponds to B_math z.
    z = x @ A
    if act is not None:
        z = act(z)

    y_lr = z @ B
    d_out = B.shape[-1]
    u = expand_k(z, d_out)
    K = math.ceil(d_out / z.shape[-1])
    y = y_lr + (alpha / math.sqrt(K)) * u
    if bias is not None:
        y = y + bias
    return y
\end{lstlisting}
\vspace{2pt}

\paragraph{Reference implementation of the fold operation.}
Folding a trained \dlr layer reduces to a single in-place update of \(B_{\mathrm{code}}\) that absorbs the structured residual; equivalently, in the paper notation, it applies \(B \leftarrow B + (\alpha/\sqrt{K})R^\top\) in \Eqref{eq:fold}.
The resulting checkpoint is a drop-in replacement for the underlying rank-$r$ decoder. The operation is exact (no learning, no approximation) and runs once at training termination.

\lstset{firstnumber=1}
\vspace{2pt}
\begin{lstlisting}[language=Python,caption={Python-style reference code for \texttt{fold()}: absorb the \dlr residual into $B$ in closed form, see~\Eqref{eq:fold}.},label={lst:dlr_fold_reference}]
import math

def fold_(B, alpha):
    r, d_out = B.shape
    K = math.ceil(d_out / r)
    scale = alpha / math.sqrt(K)
    for j in range(r):
        lo = j * K
        hi = min((j + 1) * K, d_out)
        B[j, lo:hi].add_(scale)
    return B
\end{lstlisting}
\vspace{2pt}

The loop is the transpose-layout version of the paper update \(B_{\mathrm{math}} \mathrel{+{=}} (\alpha/\sqrt{K})R^{\top}\) in \Eqref{eq:fold}.
Equivalently, for the code layout \(B_{\mathrm{code}}=B_{\mathrm{math}}^\top\), row \(j\) of \texttt{B} receives the constant \(\alpha/\sqrt{K}\) on the output coordinates in group \(\mathcal{G}_j\).
It runs in \(\mathcal{O}(d_{\mathrm{out}})\) time and \(\mathcal{O}(1)\) extra memory beyond the in-place buffer of \(B^{\star}\). After folding, the original \(\alpha\), \(K\), the \texttt{Expand}\textsubscript{\texttt{K}} branch, and any buffer storing \(R\) are discarded; what remains is a vanilla rank-\(r\) low-rank layer parameterized by \((A, B^{\star})\).

\paragraph{Fold consistency on a trained checkpoint.}
We verify the closed-form fold operation on one trained 1B \dlr+CoLA checkpoint used for this fold-consistency diagnostic.
The checkpoint loads without missing or unexpected keys, and folding updates 168 \dlr layers while leaving the parameter count unchanged (609.31M before and after folding).
Under BF16 evaluation on C4 validation with the same 10M-token budget and sequence length 256 as our training-time evaluation, perplexity changes from 14.6295 before folding to 14.6301 after folding (\(\Delta\)PPL \(=+0.0006\)).
The folded and unfolded logits match within mean absolute error \(1.91\times 10^{-2}\) and safe mean relative error \(1.44\times 10^{-2}\), passing our BF16 tolerance.
Folding also removes the training-only residual branch in our single-GPU diagnostic, reducing median forward latency from 29.59 ms to 21.18 ms with no change in peak memory.
These results confirm that the fold identity in \Eqref{eq:fold} holds up to floating-point roundoff in trained checkpoints.

\section{Additional ablations and long-horizon results}
\label{appx:additional_ablation}

\subsection{Variance-preserving correction ablation}
\label{appx:varcorr_ablation}

We ablate the variance/scale correction of the duplicated residual by training \dlr+CoLA with the default scaling \(\alpha/\sqrt{K}\) versus an uncorrected variant (no \(1/\sqrt{K}\)), across model scales.
\Tabref{tab:dlr_varcorr_1b} reports validation perplexity and shows that the correction consistently improves performance, supporting the variance-preserving design used in \Eqref{eq:dlr}.

\begin{table}[h]
\centering
\vspace{0.1in}
\setlength{\tabcolsep}{4pt}
\caption{\textbf{Variance-preserving correction ablation (CoLA backbone).}
Validation perplexity across model scales for \dlr+CoLA with vs.\ without the $1/\sqrt{K}$ correction; mean$\pm$std across seeds $\{41,42,43,44,45\}$.}
\vspace{0.1in}
\label{tab:dlr_varcorr_1b}

\begin{tabular}{l|cccc}
\toprule
\dlr+ CoLA  & 60M & 130M & 350M & 1B \\
\midrule
w/ correction
& 32.96\textsubscript{\tiny{$\pm$0.06}}
& 23.80\textsubscript{\tiny{$\pm$0.40}}
& 18.38\textsubscript{\tiny{$\pm$0.03}}
& 14.26\textsubscript{\tiny{$\pm$0.02}} \\

w/o correction
& 33.80\textsubscript{\tiny{$\pm$0.14}}
& 24.46\textsubscript{\tiny{$\pm$0.05}}
& 19.02\textsubscript{\tiny{{$\pm$0.03}}}
& 14.60\textsubscript{\tiny{{$\pm$0.02}}} \\
\bottomrule
\end{tabular}
\end{table}

\subsection{Initialization-offset ablation}
\label{appx:init_offset_ablation}

From the folded perspective, \dlr adds a structured block-constant offset to the learned decoder \(B\).
To separate this static initialization effect from the active residual path during training, we compare full \dlr+CoLA with an \emph{Init-offset CoLA} variant: at initialization, we fold the same \((\alpha/\sqrt{K})R^\top\) offset into \(B\), then disable the \dlr branch and train the resulting CoLA model normally.
\Tabref{tab:init_offset_cola_1b} shows that the initialization offset is a useful structured prior, improving over CoLA, but it does not match full \dlr training.
The remaining gap indicates that the active training-time residual path provides additional optimization benefit beyond a static folded initialization of \(B\).

\begin{table}[h]
\centering
\vspace{0.1in}
\caption{\textbf{Initialization-offset ablation at LLaMA-1B.}
Validation perplexity on C4.
Init-offset CoLA folds the \dlr offset into \(B\) at initialization and then trains without the \dlr branch.}
\vspace{0.1in}
\label{tab:init_offset_cola_1b}
\begin{tabular}{lc}
\toprule
Method & PPL \(\downarrow\) \\
\midrule
CoLA & 15.76 \\
Init-offset CoLA & 14.79\textsubscript{\tiny{$\pm$0.04}} \\
\dlr+CoLA & 14.26\textsubscript{\tiny{$\pm$0.02}} \\
\bottomrule
\end{tabular}
\end{table}

\subsection{Target-rank sensitivity}
\label{appx:rank_sensitivity}

To probe sensitivity to the target rank, we train \dlr+CoLA models with reduced ranks \(r\in\{0.5r_0,\,0.75r_0\}\), where \(r_0\) is the default rank used throughout the paper (\Tabref{tab:train_hparams}).
All settings follow the main recipe and use five random seeds \(\{41,42,43,44,45\}\).
\Tabref{tab:dlr_rank_sensitivity} shows a smooth, monotonic trade-off: decreasing \(r\) degrades perplexity gradually rather than catastrophically.
For instance, at 1B scale, reducing the rank from \(r_0=512\) to \(384\) (0.75\(r_0\)) increases PPL from 14.26 to 14.77, and further to 15.67 at \(256\) (0.5\(r_0\)).
Similar trends hold at 350M (18.38 \(\rightarrow\) 19.18 \(\rightarrow\) 20.59) and 130M (23.80 \(\rightarrow\) 24.57 \(\rightarrow\) 26.18), indicating that \dlr remains effective under tighter rank budgets.

\begin{table}[t]
\centering
\vspace{0.1in} 
\caption{\small
Rank sensitivity for \dlr+CoLA. Target rank \(r\) is varied as a fraction of the default rank \(r_0\) at each scale.
We report mean$\pm$std validation perplexity over seeds \(\{41,42,43,44,45\}\) on C4.
}
\vspace{0.1in} 
\label{tab:dlr_rank_sensitivity}

\renewcommand{\arraystretch}{1.12}
\setlength{\tabcolsep}{4.5pt}
\small
\begin{tabular}{c|ccc}
\toprule
Model params & Rank ratio & \(r\) & PPL \(\downarrow\) \\
\midrule
60M  & 0.50 & 64  & 37.26\textsubscript{\tiny{$\pm$0.12}} \\
60M  & 0.75 & 96  & 34.98\textsubscript{\tiny{$\pm$0.08}} \\
60M  & 1.00 & 128 & 32.96\textsubscript{\tiny{$\pm$0.06}} \\
\midrule
130M & 0.50 & 128 & 26.18\textsubscript{\tiny{$\pm$0.11}} \\
130M & 0.75 & 192 & 24.57\textsubscript{\tiny{$\pm$0.22}} \\
130M & 1.00 & 256 & 23.80\textsubscript{\tiny{$\pm$0.40}} \\
\midrule
350M & 0.50 & 128 & 20.59\textsubscript{\tiny{$\pm$0.01}} \\
350M & 0.75 & 192 & 19.18\textsubscript{\tiny{$\pm$0.03}} \\
350M & 1.00 & 256 & 18.38\textsubscript{\tiny{$\pm$0.03}} \\
\midrule
1B   & 0.50 & 256 & 15.67\textsubscript{\tiny{$\pm$0.02}} \\
1B   & 0.75 & 384 & 14.77\textsubscript{\tiny{$\pm$0.03}} \\
1B   & 1.00 & 512 & 14.26\textsubscript{\tiny{$\pm$0.02}} \\
\bottomrule
\end{tabular}
\end{table}

\subsection{Scaling-law token budget at 1B}
\label{appx:scaling_1b_26b}

\begin{figure}[t]
    \centering
    \includegraphics[width=.6\linewidth]{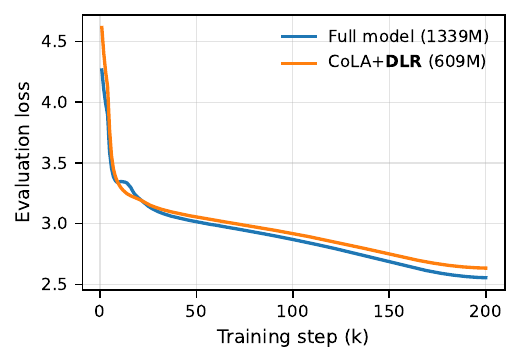}
    \caption{\textbf{Scaling-law token budget at LLaMA-1B with 26B tokens.} Validation loss trajectories on C4 for LLaMA-1B trained for 26B tokens, comparing full-rank pre-training to a low-rank backbone equipped with \dlr (r=512). The full-rank model converges to a slightly lower loss, consistent with the 1B perplexity ordering reported in \Tabref{tab:results}.}
    \label{fig:eval_loss_curve_1b_26b}
\end{figure}

To complement the final validation perplexities in \Tabref{tab:results}, we plot the full training trajectory for the 1B setting at a scaling-law token budget (26B tokens) in \Figref{fig:eval_loss_curve_1b_26b}.
The low-rank backbone equipped with \dlr closely tracks the full-rank baseline throughout training and exhibits a small but consistent gap at convergence, matching the relative performance reported in \Tabref{tab:results}.

\section{Gradient-path analysis results}
\label{app:gradlog}

We instrument the training loop to probe the gradient paths induced by the low-rank decoder and the DLR expansion operator at selected update steps.
For each probed module output $y\in\mathbb{R}^{d_{\text{out}}}$, we capture $y$ via a forward hook and compute $g_y=\partial\mathcal{L}/\partial y$ using \texttt{torch.autograd.grad}, avoiding any modification of parameter gradients.
We then form the two latent-gradient components $B^\top g_y$ and $\beta R g_y$ (with $\beta=\alpha/\sqrt{K}$), where $R$ corresponds to the DLR ``expand-then-sum'' operator (implemented as a scatter-add over duplicated groups).
Following our implementation, we reshape $g_y$ to $[N,d_{\text{out}}]$ by flattening batch and sequence dimensions, compute token-wise $\ell_2$ norms, and report the token-average norms
$\|B^\top g_y\|$ and $\|\beta R g_y\|$.
We log $\rho(t)=\|\beta R g_y\|/\|B^\top g_y\|$ and the token-average cosine similarity
$\cos_t=\langle B^\top g_y,\beta R g_y\rangle/(\|B^\top g_y\|\cdot\|\beta R g_y\|)$.
Note that $\cos_t$ measures the \emph{directional alignment of the two gradient components} (not the cosine between the parameter matrices $B$ and $R$).

\begin{figure}[t]
  \centering
  \begin{subfigure}[t]{0.48\linewidth}
    \centering
    \includegraphics[width=\linewidth]{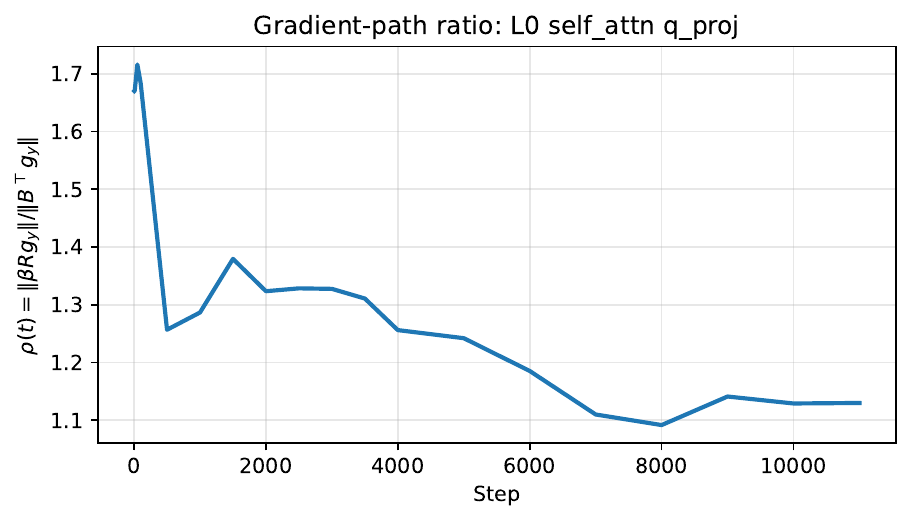}
    \caption{Ratio $\rho(t)=\|\beta R g_y\|/\|B^\top g_y\|$}
    \label{fig:gradpath_l0_q_ratio}
  \end{subfigure}\hfill
  \begin{subfigure}[t]{0.48\linewidth}
    \centering
    \includegraphics[width=\linewidth]{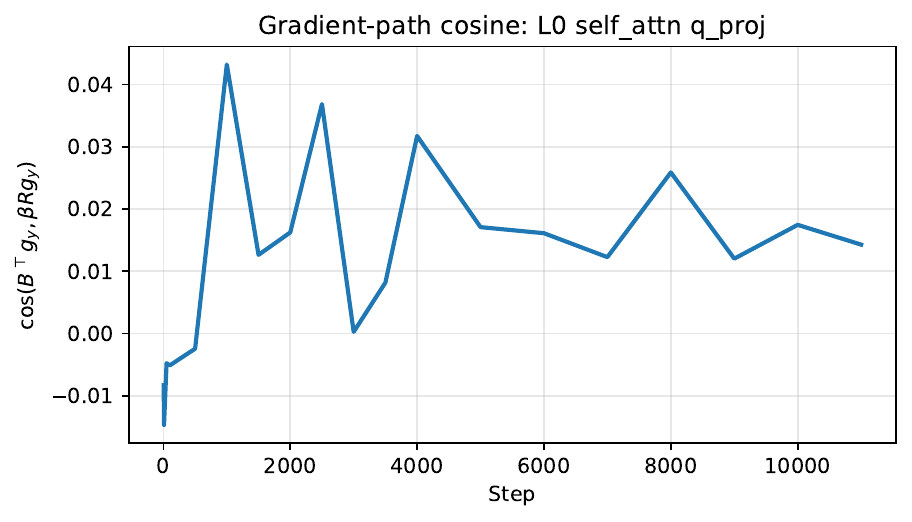}
    \caption{Cosine $\cos_t=\cos(B^\top g_y,\beta R g_y)$}
    \label{fig:gradpath_l0_q_cos}
  \end{subfigure}
  \caption{\textbf{Gradient-path diagnostics (LLaMA-350M): \texttt{L0 self\_attn q\_proj}.}
  The DLR-induced component dominates early updates ($\rho(t)>1$) and decays below $1$ as training progresses, while remaining nearly orthogonal to the baseline low-rank gradient ($\cos_t\approx 0$).}
  \label{fig:gradpath_l0_q}
\end{figure}

\begin{figure}[t]
  \centering
  \begin{subfigure}[t]{0.48\linewidth}
    \centering
    \includegraphics[width=\linewidth]{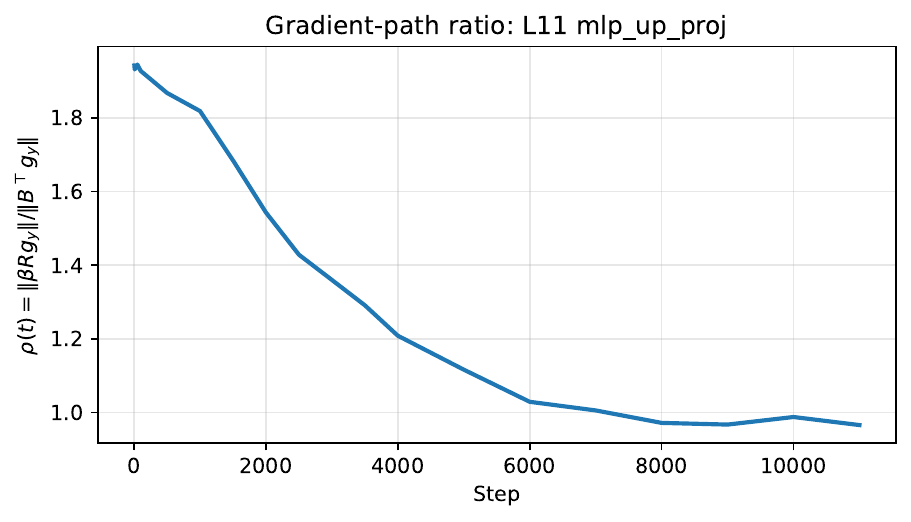}
    \caption{Ratio $\rho(t)=\|\beta R g_y\|/\|B^\top g_y\|$}
    \label{fig:gradpath_l11_up_ratio}
  \end{subfigure}\hfill
  \begin{subfigure}[t]{0.48\linewidth}
    \centering
    \includegraphics[width=\linewidth]{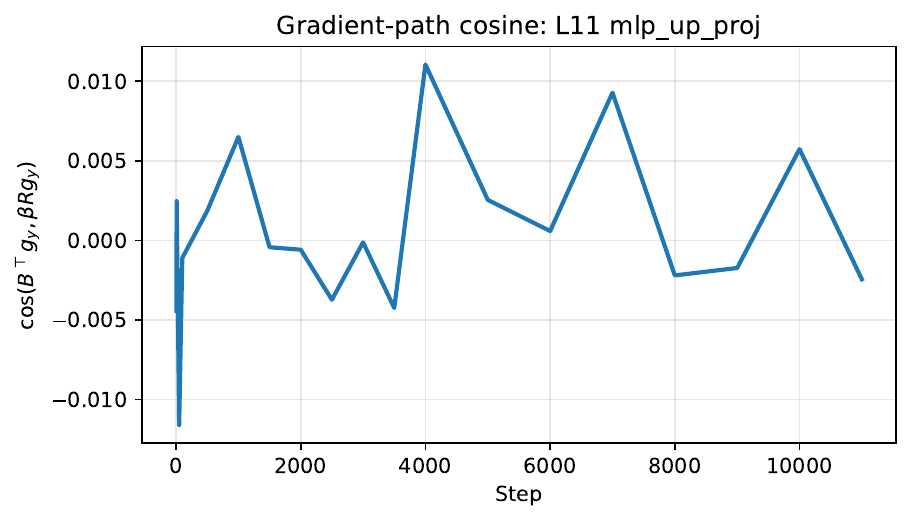}
    \caption{Cosine $\cos_t=\cos(B^\top g_y,\beta R g_y)$}
    \label{fig:gradpath_l11_up_cos}
  \end{subfigure}
  \caption{\textbf{Gradient-path diagnostics (LLaMA-350M): \texttt{L11 mlp\_up\_proj}.}
  The ratio $\rho(t)$ decreases from above $1$ to below $1$ over training, while $\cos_t$ stays close to $0$, indicating a strong but complementary DLR gradient contribution.}
  \label{fig:gradpath_l11_up}
\end{figure}

\begin{figure}[t]
  \centering
  \begin{subfigure}[t]{0.48\linewidth}
    \centering
    \includegraphics[width=\linewidth]{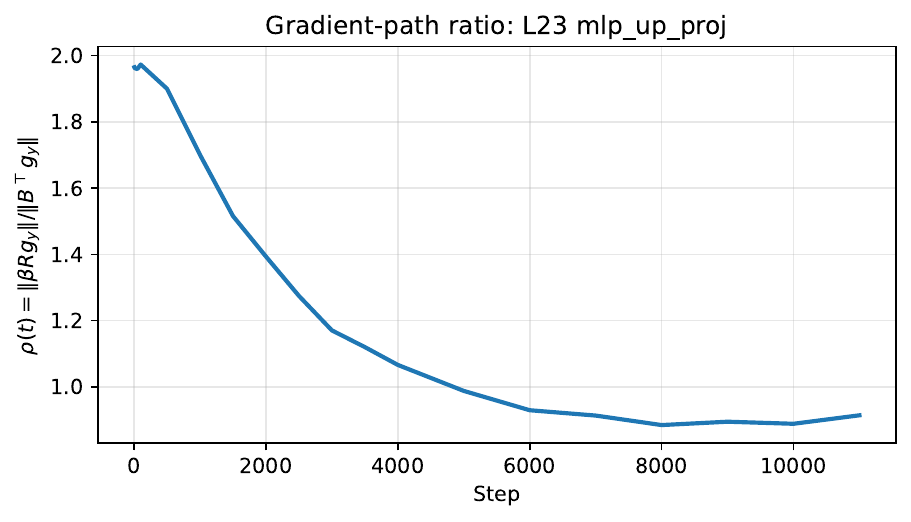}
    \caption{Ratio $\rho(t)=\|\beta R g_y\|/\|B^\top g_y\|$}
    \label{fig:gradpath_l23_up_ratio}
  \end{subfigure}\hfill
  \begin{subfigure}[t]{0.48\linewidth}
    \centering
    \includegraphics[width=\linewidth]{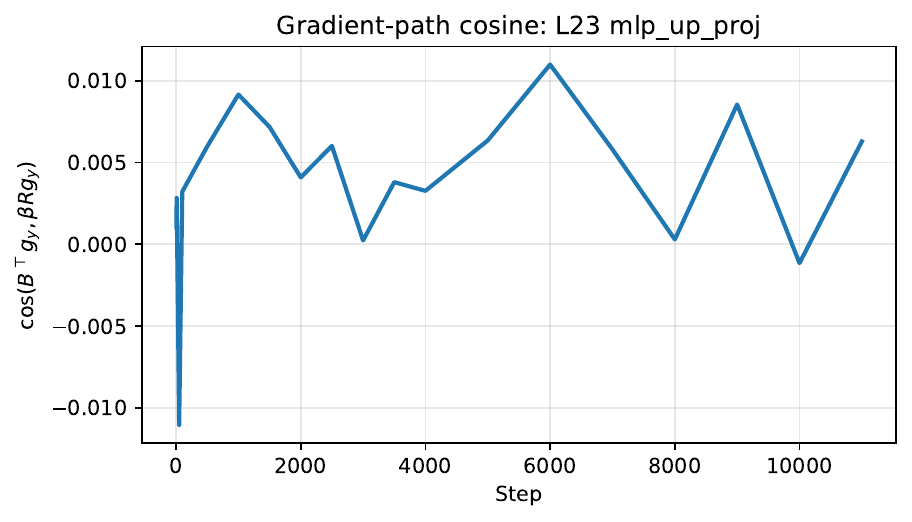}
    \caption{Cosine $\cos_t=\cos(B^\top g_y,\beta R g_y)$}
    \label{fig:gradpath_l23_up_cos}
  \end{subfigure}
  \caption{\textbf{Gradient-path diagnostics (LLaMA-350M): \texttt{L23 mlp\_up\_proj}.}
  We observe the same trend as in other layers: the DLR term dominates early ($\rho(t)>1$) and then decays, while remaining nearly orthogonal to the baseline gradient ($\cos_t\approx 0$).}
  \label{fig:gradpath_l23_up}
\end{figure}

\section{The impact of \dlr on training dynamics}
\label{app:dlr-dynamics-60m}

\begin{figure}[H]
    \centering
    \includegraphics[width=0.55\linewidth]{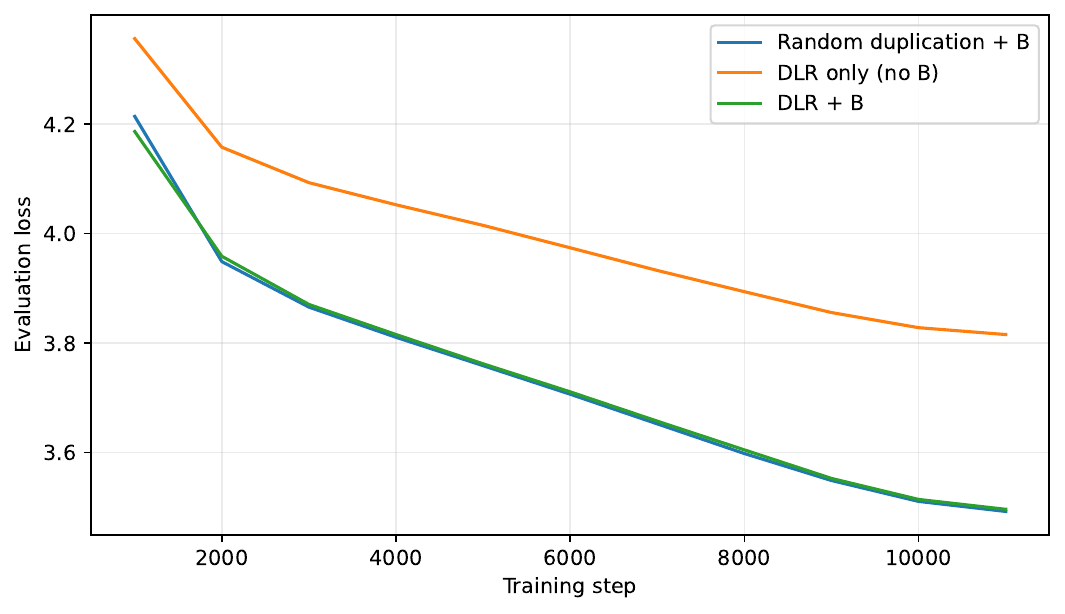}
    \caption{\textbf{Training dynamics on LLaMA-60M.}
    Evaluation loss curves for (i) \textsc{DLR}+B (default), (ii) \textsc{DLR} only (removing the learnable decoder branch \(B\)), and (iii) Random duplication + B (replacing deterministic duplication with a fixed random mapping while keeping the same variance correction \(1/\sqrt{K}\)).
    Removing \(B\) substantially degrades convergence, while \textsc{DLR}+B and Random duplication + B exhibit very similar trajectories.}
    \label{fig:llama60m-eval-loss-dlr-ablation}
\end{figure}

We provide additional training-dynamics evidence by plotting evaluation loss over pre-training for three variants:
(i) \textsc{DLR}+B (default),
(ii) \textsc{DLR} only (no learnable decoder branch \(B\)),
and (iii) Random duplication + B (fixed random duplication mapping with the same variance correction \(1/\sqrt{K}\)).
All three runs enable \texttt{torch.compile} and are executed on the same DGX machine with \(8\times\)A100 GPUs connected via NVLink.
Figure~\ref{fig:llama60m-eval-loss-dlr-ablation} shows that removing \(B\) significantly worsens optimization and yields consistently higher evaluation loss.
Meanwhile, \textsc{DLR}+B and Random duplication + B converge along nearly identical trajectories and reach comparable final losses, supporting the interpretation of DLR as an optimization/conditioning aid that complements (rather than replaces) the learnable low-rank decoder.
Despite similar convergence, Random duplication is substantially less efficient: it achieves only 507{,}186 tok/s versus 1{,}445{,}134 tok/s for \textsc{DLR}+B, because it relies on indexing-based gathers instead of contiguous duplication.
These results validate the design constraints used in the main method.
First, keeping the learnable decoder \(B\) is necessary because the duplicated residual is not intended to replace the low-rank up-projection.
Second, the residual map should be fixed and linear in \(z\): any such map, including the fixed random mapping tested here, can in principle be folded into a same-shape decoder, whereas data-dependent or nonlinear maps generally cannot.
Finally, using contiguous \texttt{repeat\_interleave} rather than indexed gathers is what lets \texttt{torch.compile} fuse the duplication into surrounding computation and recover the throughput profile of vanilla low-rank training.

\textbf{Definition of random duplication.}
For each module, we sample a \emph{fixed} index vector \(\texttt{idx}\in\{0,\dots,r-1\}^{d_{\text{out}}}\) once and reuse it throughout training, where \(\texttt{idx}\sim \texttt{torch.randint}(0,r,(d_{\text{out}},))\) under a deterministic per-module seed (base seed XOR \(\mathrm{MD5}(\texttt{module\_name})\)).
We form the expanded residual by indexing \(z_{\text{exp}}[j]=z[\texttt{idx}[j]]\) (i.e., \texttt{gather}/\texttt{index\_select}), which incurs irregular memory access and poorer fusion than deterministic \texttt{repeat\_interleave}.

\section{Supervised fine-tuning details}
\label{appx:sft_details}

For the downstream SFT evaluation in \Tabref{tab:sft}, we fine-tune the 1B Full-Rank, CoLA, and folded \dlr+CoLA checkpoints on Alpaca-cleaned (52K instructions) for 3 epochs using full-parameter SFT.
All SFT runs use 16$\times$H100 DDP, effective batch size 64, learning rate \(2{\times}10^{-5}\) with cosine schedule, BF16, and maximum sequence length 512.
We then evaluate ARC-Challenge, BoolQ, HellaSwag, MMLU, PIQA, and WinoGrande with \texttt{lm-evaluation-harness} in the zero-shot setting.
For each method, the three SFT seeds are applied to the same fixed pre-trained checkpoint, so the reported standard deviations capture SFT-side data-ordering variability rather than pre-training initialization variance.
BoolQ and MMLU collapse to constant chance-level accuracy with zero across-seed variance for all three checkpoints, consistent with the behavior of pre-training-only 1B models on these formats~\citep{sltrain,glentis2025scalableparametermemoryefficient}; we therefore interpret the average mainly through the tasks where the checkpoints differentiate.

\section{Zero-shot evaluation}
\label{appx:zero_shot_eval}

In addition to validation perplexity on C4, we perform zero-shot evaluations of 1B-scale models on ARC-Challenge (ARC-C), BoolQ, HellaSwag, MMLU, PIQA, and WinoGrande, using standard zero-shot prompts (no instruction tuning).
All low-rank variants use rank $r=512$ and are trained with five random seeds $\{41,42,43,44,45\}$; we report mean$\pm$std across seeds.
As shown in \Tabref{tab:zeroshot_1b}, \dlr improves the average score over the pure low-rank baseline and largely closes the gap to full-rank; we treat these results as supplementary since several tasks remain near-chance at this pre-training-only stage.

\begin{table}[h]
\centering
\vspace{0.1in}
\caption{\textbf{Zero-shot evaluation on 1B models.}
Each cell reports zero-shot accuracy as mean$\pm$std across five pre-training seeds $\{41,42,43,44,45\}$;
the Avg.\ column is the across-task mean of the per-task means.
BoolQ and MMLU saturate at constant chance-level accuracy (std$=0$) for all four checkpoints,
consistent with the well-documented behavior of pre-training-only 1B models on these formats.}
\label{tab:zeroshot_1b}
\vspace{0.05in}
\small
\setlength{\tabcolsep}{4pt}
\resizebox{\textwidth}{!}{%
\begin{tabular}{l|cccccc|c}
\toprule
Method (1B, $r{=}512$) & ARC-C & BoolQ & HellaSwag & MMLU & PIQA & WinoGrande & Avg. \\
\midrule
Full-Rank      & 22.06$\pm$0.57 & 37.83$\pm$0.00 & 34.94$\pm$0.18 & 22.95$\pm$0.00 & 67.57$\pm$0.63 & 51.10$\pm$1.13 & 39.41 \\
Low-Rank       & 19.78$\pm$0.76 & 37.83$\pm$0.00 & 27.47$\pm$0.16 & 22.95$\pm$0.00 & 60.28$\pm$0.76 & 51.16$\pm$1.21 & 36.58 \\
\dlr+Low-Rank  & 21.59$\pm$0.65 & 37.83$\pm$0.00 & 32.52$\pm$1.03 & 22.95$\pm$0.00 & 65.59$\pm$0.68 & 50.69$\pm$0.24 & 38.53 \\
\dlr+CoLA      & 22.29$\pm$0.57 & 37.83$\pm$0.00 & 34.20$\pm$0.24 & 22.95$\pm$0.00 & 66.81$\pm$0.58 & 50.61$\pm$1.57 & 39.12 \\
\bottomrule
\end{tabular}%
}
\end{table}

\section{Related work}
\label{appx:related}

Related work on efficient pre-training spans four complementary directions. First, low-rank adapters popularized by LoRA learn update-level low-rank corrections \citep{lora}; pre-training variants such as ReLoRA and SwitchLoRA recover higher effective ranks via periodic merges or frequent subspace switching \citep{relora,switchlora}, while LORO directly optimizes a fixed-rank factorization on the low-rank manifold \citep{loro}. Second, memory-efficient gradient projection methods (GaLore and variants) project gradients into low-rank subspaces to reduce optimizer state, with Fira stabilizing training via norm-based scaling and a norm-growth limiter \citep{gaLore,qgalore,fira}. Third, sparse-plus-low-rank designs increase expressivity by adding sparse components: SLTrain employs a fixed unstructured support with low memory overhead \citep{sltrain,glentis2025scalableparametermemoryefficient}, LOST co-designs complementary low-rank and channel-wise structured sparse components guided by an SVD initialization \citep{lost}. A closely related latent-shaping line improves low-rank trainability without introducing an explicit sparse pathway: CoLA inserts a nonlinearity between factors to enforce low-rank structure in activations \citep{cola}. LaX \citep{LaX} designs inter-layer latent crossing to enhance the capacity of low-rank models by establishing information flow across low-rank spaces of different layers. LoR2C \citep{LoR2C} introduces low-rank residual connections within the model layers to solve gradient vanishing during LoRA fine-tuning. ResLoRA \citep{ResLoRA} integrates residual paths into the LoRA fine-tuning framework to accelerate training convergence and improve performance, utilizing merging strategies to ensure no additional computational cost during inference. Similarly, FST~\citep{hu2024acceleratingtransformerpretraining24} and SLoPe~\citep{mozaffari2025slopedoubleprunedsparseplus} optimize training for semi-structured (e.g., 2:4) or block-sparse patterns to leverage specialized kernels. Finally, post-training model folding clusters redundant channels/heads and merges them with variance-preserving corrections to compress models without data or fine-tuning \citep{wang2025modelfolding,saukh2026cut}. DLR brings the folding insight \emph{into training}: it introduces a \emph{foldable} latent-space compensation within each low-rank (or CoLA-style) layer. Concretely, with a rank-$r$ latent $z=s\,\phi(A^\top x)$, DLR augments the decoder output by a structured duplicated residual, which is a fixed replication matrix that maps each latent coordinate to a small group of output channels ; compared to the above, this design reduces FLOPs and optimizer state like low-rank methods, avoids sparse-kernel overheads, and empirically delivers favorable perplexity--efficiency trade-offs across LLaMA scales under matched token budgets. Orthogonal to these four directions, a complementary line of work targets the activation-memory bottleneck by compressing activations stored for backpropagation: VeLoRA compresses intermediate activations via rank-1 sub-token projections and reconstructs them approximately during the backward pass \citep{VeLoRA}, CompAct stores low-rank, randomly projected activations and only decompresses gradients for the optimizer update, achieving $25$--$30\%$ peak-memory savings for LLaMA pre-training and up to $50\%$ for RoBERTa fine-tuning \citep{CompAct}, and tensor-decomposition methods compress activation maps using (high-order) SVD with theoretical guarantees on gradient approximation error \citep{nguyen2024activationmapcompressiontensor}. 

\Tabref{tab:related_residual_methods} is not intended as a complete taxonomy of efficient pre-training.
Instead, it isolates residual-style interventions that are easy to conflate with \dlr.
ResNet~\citep{heresnet} is included only as a conceptual anchor for identity residuals, not as an efficient pre-training baseline.
Within this narrower comparison, \dlr differs by combining a pre-training setting, a parameter-free intra-layer latent residual, and a closed-form fold into the deployed low-rank graph.

\begin{table}[t]
\centering
\small
\caption{\textbf{Where \dlr sits among residual-style methods.}
We anchor against ResNet to remind readers that residual primitives predate efficient pre-training,
then compare \dlr with three pre-training / fine-tuning residual baselines.
\dlr is the only entry that combines
(i)~pre-training applicability,
(ii)~a parameter-free residual path, and
(iii)~an exact closed-form merge into the deployment graph (\Eqref{eq:fold}).}
\label{tab:related_residual_methods}

\setlength{\tabcolsep}{5pt}
\renewcommand{\arraystretch}{1.15}
\begin{tabularx}{\textwidth}{l c >{\raggedright\arraybackslash}X c >{\raggedright\arraybackslash}X c}
\toprule
Method
& Setting
& Residual space
& Learnable?
& Main role
& Foldable? \\
\midrule

ResNet
& Full
& Activation
& No
& Stabilize optimization
& N/A (identity) \\

ResLoRA
& FT
& Adapter
& Yes
& Improve adaptation
& Sometimes \\

LoR2C
& FT
& Correction
& Yes
& Improve capacity
& Sometimes \\

LaX
& Pretrain
& Inter-layer latent
& Partial
& Cross-layer flow
& No \\

\textbf{\dlr}
& Pretrain
& Intra-layer latent
& No
& Decoder-independent gradient path
& \textbf{Yes} \\

\bottomrule
\end{tabularx}
\end{table}

\section{Impact statement and existing assets}
\label{appx:impact}

\paragraph{Impact statement.}
This work studies an efficiency technique for low-rank language-model pre-training.
Its intended positive impact is to reduce the compute and memory required to train and deploy low-rank models, which may make pre-training experiments more accessible and reduce resource use.
The method does not introduce a new application, dataset, or user-facing system, but improvements in pre-training efficiency could also lower the cost of training models that may be misused if deployed without appropriate safety evaluation.
We therefore view \dlr as a general-purpose training method whose downstream risks are primarily those of the models and deployment settings to which it is applied.

\paragraph{Existing assets and licenses.}
We use publicly available research assets and cite their original sources where they are introduced in the paper.
Pre-training uses C4~\citep{raffel2020exploring} accessed through the Hugging Face Datasets interface; tokenization uses the T5-base tokenizer~\citep{raffel2023t5}; supervised fine-tuning uses Alpaca-cleaned; and downstream evaluation uses \texttt{lm-evaluation-harness}.
We do not redistribute the raw datasets or third-party model checkpoints as part of this source package.
Users reproducing the experiments should obtain these assets from their original providers and comply with the licenses and terms listed on the corresponding dataset, model, and software package pages.
Our code release is adapted from the open-source implementation of \citet{glentis2025scalableparametermemoryefficient}; the release will include attribution and license information for the reused code and experiment scripts.

\section{Use of Large Language Models}
\label{appx:use-of-llm}
We primarily used \texttt{ChatGPT}\footnote{https://chatgpt.com/} to correct grammatical errors in the manuscript and to fix minor compilation issues in Overleaf. In addition, \texttt{Cursor}\footnote{https://cursor.com/} with \texttt{GPT-5.2-High}\footnote{https://openai.com/index/introducing-gpt-5/} was employed to debug programming errors encountered during implementation. Apart from these auxiliary uses, the research ideas, theoretical contributions, and the writing of this paper were entirely carried out by the authors. The code implementation is adapted from the open-source code of ~\citep{glentis2025scalableparametermemoryefficient}.

\end{document}